\documentclass{spie}

\usepackage{cite}

\usepackage{times} 
\usepackage{indentfirst} 

\usepackage{amsmath}
\usepackage{amsfonts}
\usepackage{multirow}
\usepackage{mathtools}
\usepackage{xcolor}
\usepackage{doi}
\usepackage{xkeyval}

\DeclareMathOperator{\E}{E}
\newcommand{\eqdef}{\overset{\mathrm{def}}{=}}

\title{Next integrated result modelling for stopping the text field recognition process in a video using a result model with per-character alternatives}

\author{Konstantin Bulatov\supit{1,2,3}, Boris Savelyev\supit{2}, Vladimir V. Arlazarov\supit{2,3}
  \skiplinehalf
  \normalsize 
  \supit{1} Moscow Institute of Physics and Technology ``MIPT'', Moscow, Russia; \\
  \supit{2} Smart Engines, Moscow, Russia; \\
  \supit{3} Federal Research Center ``Computer Science and Control'' of Russian Academy of Sciences, Moscow, Russia
}

\begin{document}

\maketitle

\begin{abstract}
    In the field of document analysis and recognition using mobile devices for capturing, and the field of object recognition in a video stream, an important problem is determining the time when the capturing process should be stopped. Efficient stopping influences not only the total time spent for performing recognition and data entry, but the expected accuracy of the result as well. This paper is directed on extending the stopping method based on next integrated recognition result modelling, in order for it to be used within a string result recognition model with per-character alternatives. The stopping method and notes on its extension are described, and experimental evaluation is performed on an open dataset MIDV-500. The method was compares with previously published methods based on input observations clustering. The obtained results indicate that the stopping method based on the next integrated result modelling allows to achieve higher accuracy, even when compared with the best achievable configuration of the competing methods.

\end{abstract}

\section{Introduction}

Modern document entry systems allow to automatize the process of data extraction from various documents, either business, regulatory, or personal. Such systems are used for creating digital archives of historical documents \cite{VanPhan2016}, recognition of small-scale documents such as business cards \cite{google-doc-2}, ID documents, driving licences, passports \cite{midv500-arxiv}, as well as large-scale business documents \cite{google-doc-3}.

\begin{figure}[ht!]
  \centering
  \includegraphics[width=0.65\textwidth]{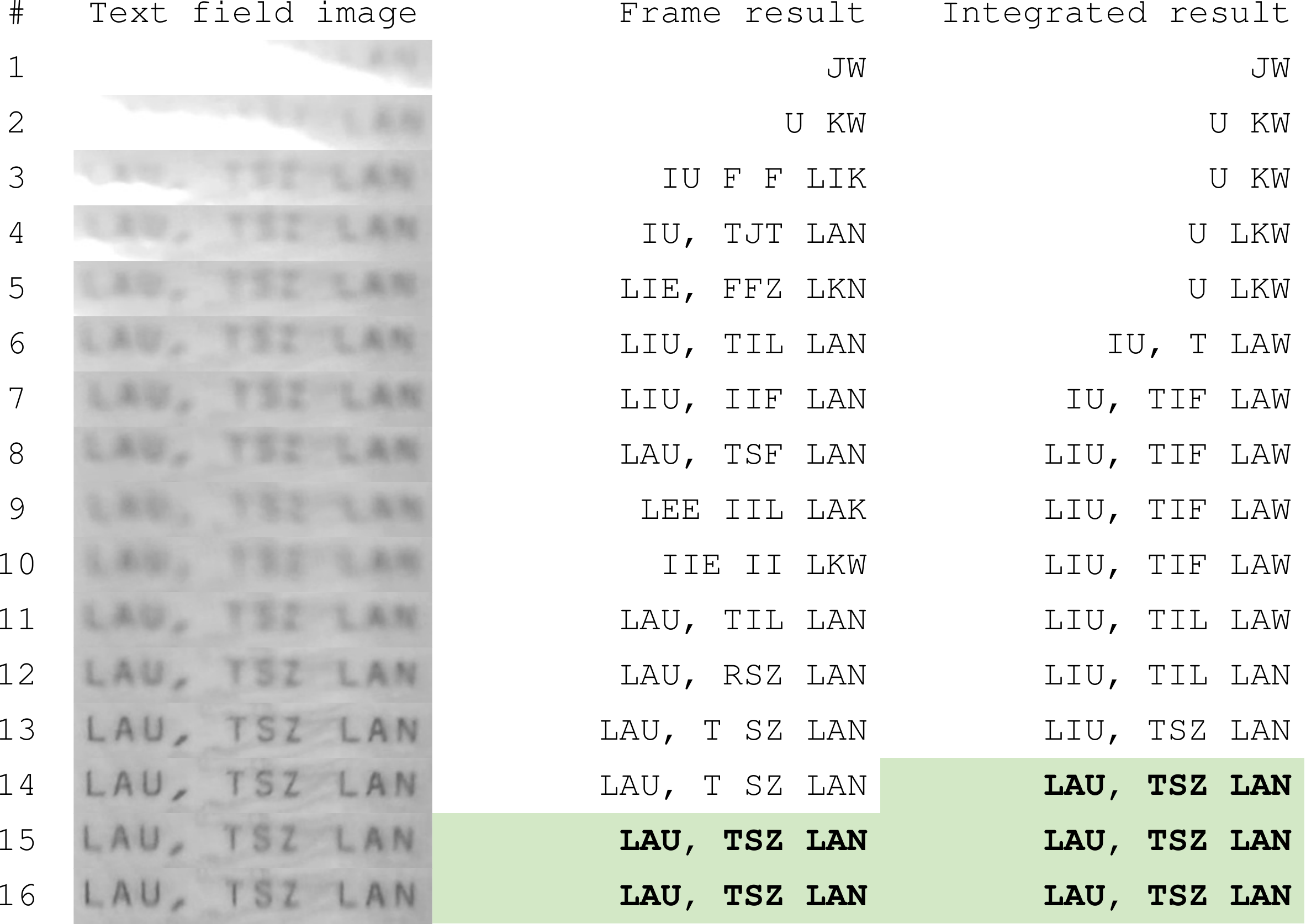}
  \caption{Example of per-frame recognition results combination in a video stream. Correct recognition results are highlighted. Images are taken from MIDV-500 dataset \cite{midv500-arxiv}}
  \label{fig:integration_illustration}
\end{figure}

Increasing computational power of mobile devices and rising technical characteristics of small-scale digital cameras lead to increased interest in methods for automatic document entry using mobile devices \cite{google-doc-4, google-doc-10, Povolotskiy_Tropin, Shemiakina, Chernyshova}. As a rule, regular smartphones are used for document recognition, due to relatively low cost, sufficient computational power for performing recognition tasks, and ability of capturing video (or sequence of images). The ability to capture video is one of the most important advantages over traditional scanners, as in such case more information could be retrieved in comparison with a single image, and each newly acquired document image may be used to improve the recognition result \cite{SMART_IDREADER_ICDAR}. Figure \ref{fig:integration_illustration} illustrates an example of per-frame recognition results combination in a video stream. As it can be seen, the correct integrated result may be acquired even before any individual frame result is correct.

While processing the sequence of frames and combining the per-frame recognition results a single more precise one, the problem arises -- when this process should terminate? The capturing process in a general case might not be naturally limited, and if a sufficiently good combination strategy is employed the increase of the number of integrated observations the expected result precision also increases \cite{vestnik_integration}. However the time required to perform recognition and output the final result is also very important and thus efficient strategies for video stream recognition stopping  should be developed and further studied. 

The optimal stopping problems themselves occupy a special place in mathematical statistics and decision theory \cite{FERGUSON, The-Monotone-Case-Approach}. Some methods were also proposed for the video stream recognition problem \cite{stopper_slavin, Bulatov2019}. In \cite{stopper_slavin} a method was presented which consisted of clustering the set of per-frame field recognition results, estimating a confidence score for each cluster, and making a stopping decision based on three parameters: cluster size, cluster confidence, and the total number of processed observations. The method can be applied in two ways: the clusters may be formed from the initial per-frame recognition results, or from the integrated results obtained on each stage. This method, however, is not fully formalized and raises the questions of tuning the clustering parameters. In \cite{Bulatov2019} a method is proposed, which considers the video stream recognition stopping as a monotone sequential decision problem. It presents a stopping strategy derived from the properties of monotone stopping problems, however it was tested only for text recognition results as simple strings, without any per-character alternatives. At the same time an extended string recognition result model with per-character alternatives is an important way of text recognition result representation: it is used for recognition results post-processing \cite{LLOBET} and was shown to be valuable for improving the integrated result precision \cite{vestnik_integration}.

The goal of this paper is to investigate the applicability of stopping strategy introduced in \cite{Bulatov2019} for text string recognition with per-character alternatives and to compare it with alternative methods, which are already adapted for such recognition result model. In section \ref{sec:framework} a brief description of the stopping method is given, and in section \ref{sec:evaluation} experimental evaluation and comparison for stopping methods is presented. 

\section{Method description} \label{sec:framework}

In order to provide a description of the stopping method, let us consider text string recognition in a video stream as a sequential decision problem. Let $\mathbb{X}$ represent a set of all possible text string recognition results, and the task is to recognize a text string with correct value $x^* \in \mathbb{X}$ given a sequence of images $I_1, I_2, I_3, \ldots$ which are obtained one at a time. At stage $n$ the image $I_n$ is recognized and the per-frame recognition result $x_n\in\mathbb{X}$ is obtained. After $x_n$ is obtained the results $x_1, x_2, \ldots, x_n$ are combined using some combination algorithm to produce an integrated result $R_n \in \mathbb{X}$. The stopping decision is now to either stop the process and use $R_n$ as the final recognition result, or continue the process in an effort to obtain in the future the integrated result with higher expected accuracy. If the process is stopped at stage $n$ the penalty is paid in form of a linear combination of distance from the obtained result to the correct one (a ``price for error'') and the number of frames process (a ``price for time''):
\begin{equation}
    \label{eq:loss_function}
    L_n = \rho(R_n, x^*) + c \cdot n,
\end{equation}
where $\rho$ is a metric function on the set $\mathbb{X}$, and $c$ is a constant representing the price paid for each observation.

The stopping rule can be formally expressed as a random variable $N$ (the stopping time), and the stopping method defines a distribution which the variable $N$ takes given the observations $x_1, x_2, x_3, \ldots$. The stopping problem is an optimization problem with a goal to minimize the expected loss, which can be expressed as follows:
\begin{equation}
    \label{eq:stopping_problem}
    \E(L_N(X_1, X_2, \ldots, X_N)) \rightarrow \min\limits_{N},
\end{equation}
where $\E(\cdot)$ is a mathematical expectation, and $X_1, X_2, \ldots, X_k$ are random recognition results with identical joint distribution with $x^*$ of which $x_1, x_2, \ldots, x_k$ are realizations observed at stages $1, 2, \ldots, k$.

The stopping method proposed in \cite{Bulatov2019} is relying on an assumption that the expected distances between two consecutive integrated results decrease over time:
\begin{equation}
    \label{eq:diminishing_returns_assumption}
    \E(\rho(R_n, R_{n+1})) \ge \E(\rho(R_{n+1}, R_{n+2})), \quad \forall n > 0,
\end{equation}
which allows to consider the text field recognition problem in a video stream as a monotone stopping problem. In monotone stopping problems if at some stage $n$ the loss function is not higher that the expected loss at the next stage, then this will be true for all later stages as well. For monotone stopping problems with finite horizon the optimal stopping rule is myopic, i.e. the one calling for stopping at stage $n$ if the current loss is not higher than the loss which will be suffered if the process is stopped at stage $n+1$. 

Since $x^*$ is unknown at the moment of making the stopping decision (and thus the loss function \eqref{eq:loss_function} cannot be computed), in \cite{Bulatov2019} it is proposed to use triangle inequality and threshold the upper boundary for the loss difference, by estimating the expected distance between the current integrated result and the next one. To achieve this, a modelling of the next integrated result is proposed, defining the following stopping rule:
\begin{equation}
    \label{eq:stopping_rule}
    \text{Stop if } \hat{\Delta}_n \le c, \qquad \hat{\Delta}_n \eqdef \cfrac{1}{n+1}\left(\delta + \sum\limits_{i=1}^n \rho(R_n, R(x_1, x_2, \ldots, x_n, x_i))\right),
\end{equation}
where $c$ is an observation cost (essentially, a threshold parameter of the stopping method), $\delta$ is an external parameter, and $R(x_1, x_2, \ldots, x_n, x_i)$ is a modeled integration result of all consecutive observations obtained by the stage $n$ concatenated with $i$-th observation.

It is noted in \cite{Bulatov2019} that the concrete method of modelling the next integrated result might depend on the nature of the combination algorithm and other specifics of the problem, however the proposed method could still be used in a quite general case, by replacing the recognition results combination method and the metric function $\rho$. In the original paper the experiments were conducted using Tesseract \cite{tesseract-paper} as the recognition algorithm, simple string of characters as a recognized string representation, a normalized Levenshtein distance \cite{ngld_yujian} as a metric function $\rho$, and ROVER \cite{FISCUS-659110} as a combination algorithm. It was not clear whether this stopping method would be effective for an extended string recognition result model, containing per-character classification alternatives. In the extended model, the string recognition result can be represented as a matrix of alternatives:
\begin{equation}
    \label{eq:extended_result}
x=\left(\begin{array}{ccc}
(q_{11}, c_{11}) & \cdots & (q_{M1}, c_{M1})\\
\vdots & \ddots & \vdots\\
(q_{1K}, c_{1K}) & \cdots & (q_{MK}, c_{MK})
\end{array}\right), \qquad \forall i \quad \sum\limits_{j=1}^K q_{ij} = 1,
\end{equation}
where $c_{ij}$ are character labels, $q_{ij}\in[0,1]$ -- class membership estimations for each character, $K$ -- the size of the alphabet, and $M$ -- the length of the string. The combination algorithms for string recognition result in this extended model can be viewed as a generalization of the ROVER approach, and to define the metric function $\rho$ a generalized Levenshtein distance may be used after defining the metric on the individual character classification results \cite{vestnik_integration}.

To compare different stopping rules the expected performance profiles can be used -- a methodology from the field of anytime algorithms \cite{Zilberstein_1996}. Expected performance profiles are graphical plots which show dependence of the expected accuracy on the expected time required to obtain it.

\section{Evaluation} \label{sec:evaluation}

In order to evaluate the stopping method described in section \ref{sec:framework} we used an open dataset MIDV-500 \cite{midv500-arxiv} which contains 500 video sequences of 50 types of identity documents with ground truth. Each original clip contained 30 frames. The frames on which the document was not fully visible were removed from the consideration, and the resulting clip was repeated in a loop until the original size of 30 frames was reached. 

The ground truth in the MIDV-500 contains both ideal values for text field recognition and the ideal geometric coordinates, i.e. for each field its geometric position in the document boundaries is known, making it possible to crop the field from any frame of the dataset. Text fields were cropped with margins with width equal to 30\% of the smallest text field bounding box side. Since physical dimensions of each document type in the MIDV-500 dataset is also known, it is possible to crop each field in a uniform resolution. For recognition, all text fields were cropped with the resolution of 300 DPI. After cropping each text field was recognized using a text string recognition subsystem of Smart IDReader document recognition software \cite{SMART_IDREADER_ICDAR}, obtaining the recognized value as a sequence of character classification results with alternatives. For combination of per-frame recognition result a method from \cite{vestnik_integration} was used, which could be regarded as a generalization of the ROVER \cite{FISCUS-659110} approach for string recognition results with per-character alternatives. As a distance metric $\rho$ a normalized version of the generalized Levenshtein distance was used, with a taxicab metric for individual character classification results.

In \cite{stopper_slavin} a stopping method was proposed, which was based on clusterization of the set of text field recognition results to $n$ clusters, and making a stopping decision based on some properties of the most populous cluster. The method proposed in \cite{Bulatov2019} and described in section \ref{sec:framework} was compared with this method in the original paper, however since the paper was focused on a simplified string recognition result model, not all features of the stopping rule presented in \cite{stopper_slavin} were used, as the per-character alternatives were not available when using Tesseract as the text string recognition algorithm.

The clusterization of the observations is performed by their lengths (i.e. by the number of characters in the obtained string recognition results). For each cluster its confidence value is computed according to the following formula:
\begin{equation}
    \label{eq:cluster_confidence}
    Q(C) = 1 - \prod\limits_{x\in C}\left(1 - \min\limits_{i=1}^{M(C)}\left\{\max\limits_{j=1}^K{q_{ij}(x)}\right\} \right),
\end{equation}
where $C$ is a cluster of observations with the same length $M(C)$. The stopping decision is made by three thresholding: the size of the largest cluster, the confidence of the largest cluster, and, if there is more than one cluster, the difference between confidences of the two largest clusters. Such thresholding meant that there are three stopping rule parameters (three thresholds). 

\begin{figure}[ht!]
  \centering
  \includegraphics[width=0.7\textwidth]{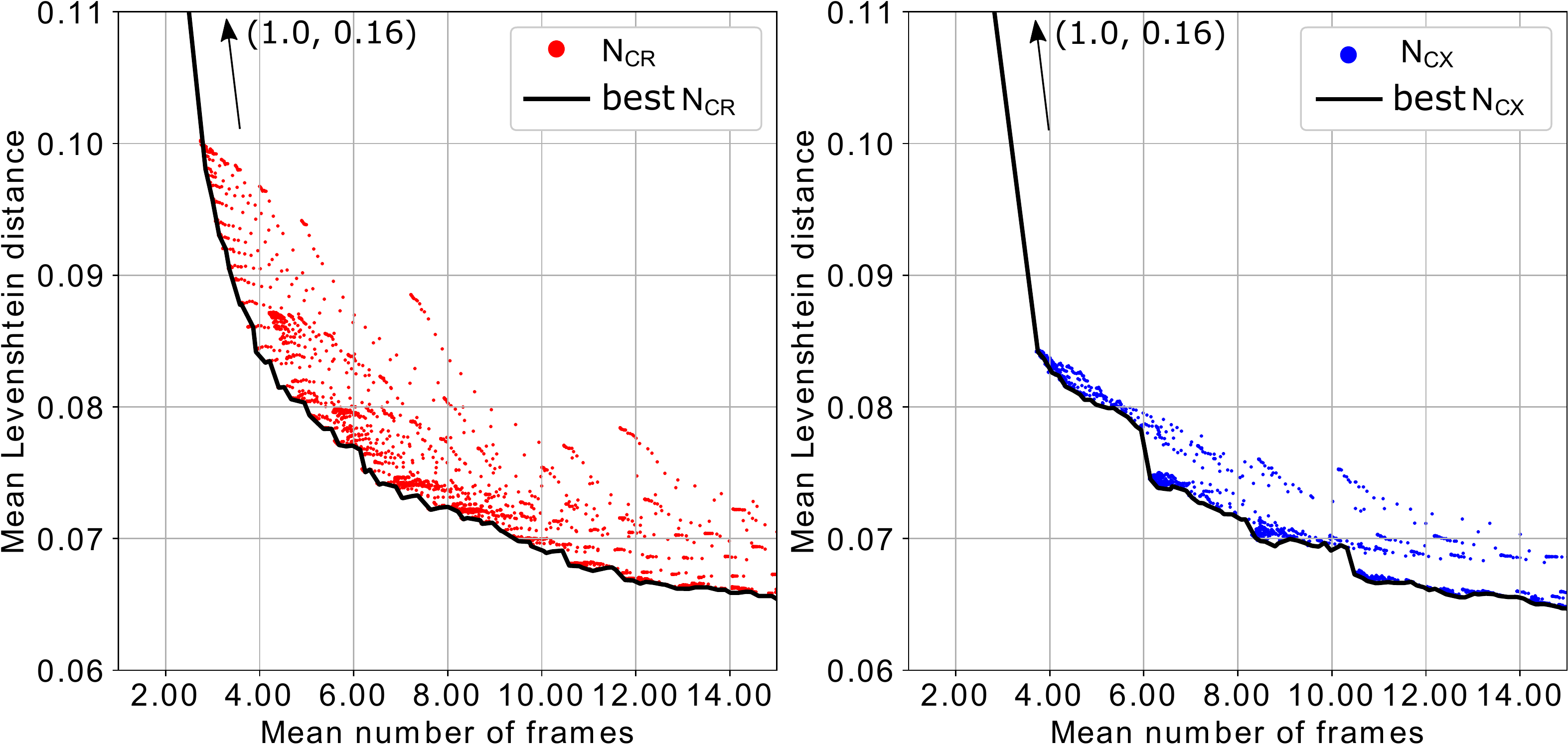}
  \caption{Quality maps for stopping method described in \cite{stopper_slavin} in two implementation variations: clusterization of integrate results (left) and of the per-frame results (right). Black line designates the best achievable result}
  \label{fig:clustering_stopper}
\end{figure}

Two variations of the stopping method proposed in \cite{stopper_slavin} can be realized -- the first, denoted hereinafter as $N_{CX}$ which treats input observations $x_1, x_2, \ldots, x_n$ as strings to compose clusters with, and the second -- $N_{CR}$ -- treats the integrated results $R_1, R_2, \ldots, R_n$ as observations and components of the clusters. Figure \ref{fig:clustering_stopper} illustrates the quality maps of the both approaches with variation of all three thresholds: each data point represent the mean number of observations processed before stopping and the mean distance of the integrated result to the correct value.

One of the main disadvantages of this stopping methods is that it is unclear how to jointly select the values for all thresholds to achieve the highest efficiency. In Figure \ref{fig:clustering_stopper} the black line represents the best option constructed a posteriori, which will be used for comparison with the method described in section \ref{sec:framework}.

\begin{figure}[ht!]
  \centering
  \includegraphics[width=0.7\textwidth]{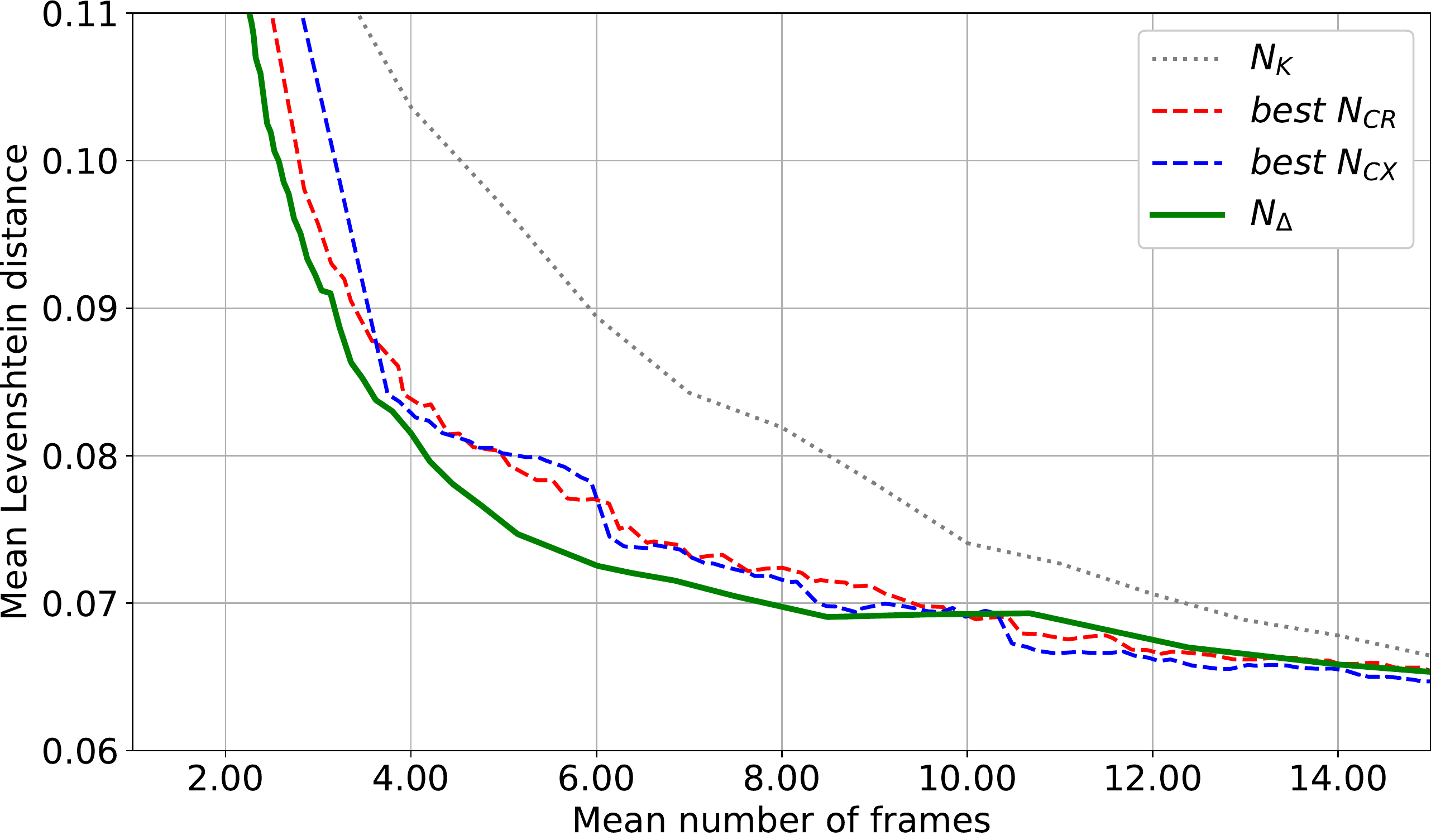}
  \caption{Expected performance profiles for the baseline stopping rule (simple integration, $N_K$), best versions of the clustering stopping rules, and the stopping rule $N_\Delta$, described in section \ref{sec:framework}}
  \label{fig:comparison}
\end{figure}

Figure \ref{fig:comparison} illustrates the expected performance profiles comparison for the best achievable versions of the stopping rules $N_{CX}$ and $N_{CR}$, the stopping rule based on the modelling of the next integrated result $N_\Delta$, described in section \ref{sec:framework}, and, as a baseline, a simple stopping rule $N_K$ which stops after observing $K$-th per-frame result. It can be seen that even though the best versions of the clustering stopping rules were evaluated, without clear understanding of how to obtain these jointly optimal threshold values, the method $N_\Delta$ still outperforms them.

\begin{table}
	\caption{Achieved values of average distance from the integrated result to the correct field value at stopping time with restricted average number of processed observations}
	\label{tbl:comparison}       
	\resizebox{\textwidth}{!}{
		\begin{tabular}{|l|l|l|l|l|l|l|l|l|}
			\hline
			\multirow{2}{*}{\parbox{0.12\columnwidth}{Stopping method}} & \multicolumn{8}{c|}{Limitation to the average number of observations} \\
			
			\cline{2-9}
			&
			\parbox{0.11\columnwidth}{$\le 3$} & 
			\parbox{0.11\columnwidth}{$\le 4$} & 
			\parbox{0.11\columnwidth}{$\le 5$} & 
			\parbox{0.11\columnwidth}{$\le 6$}  &
			\parbox{0.11\columnwidth}{$\le 7$} & 
			\parbox{0.11\columnwidth}{$\le 8$}&
			\parbox{0.11\columnwidth}{$\le 9$} & 
			\parbox{0.11\columnwidth}{$\le 10$}  

			\\
			\hline
			
			\parbox{0.14\columnwidth}{best $N_{CX}$} & $0.161$
			& $0.084$ & $0.080$ & $0.078$ & $0.074$ & $0.072$ & $\mathbf{0.069}$ & $\mathbf{0.069}$

			\\
			\hline
			
			\parbox{0.14\columnwidth}{best $N_{CR}$} & $0.096$ & $0.084$ & $0.080$ & $0.077$ & $0.074$ & $0.072$ & $0.071$ & $\mathbf{0.069}$

			\\
			\hline
			
			\parbox{0.14\columnwidth}{$N_K$} & $0.115$ & $0.104$ & $0.097$ & $0.089$ & $0.084$ & $0.082$ & $0.078$ & $0.074$

			\\
			\hline
			
			\parbox{0.14\columnwidth}{$N_\Delta$} & $\mathbf{0.092}$ & $\mathbf{0.082}$ & $\mathbf{0.076}$ & $\mathbf{0.073}$ & $\mathbf{0.071}$ & $\mathbf{0.070}$ & $\mathbf{0.069}$ & $\mathbf{0.069}$
			\\
			
			\hline
		\end{tabular}
	}
\end{table}

Table \ref{tbl:comparison} shows the achieved mean integrated result accuracy (in terms of distance to the correct value) at stopping time, using the evaluated stopping rules and with restrictions to the mean number of processed observations. It can be seen that the method based on modelling the next integrated result and thresholding the estimation of the expected distance from the current result to the next one ($N_\Delta$) outperforms the other methods. In particular, it allows to achieve higher result quality with the same average number of processed observations even when compared with the best achievable version of the previously proposed method \cite{stopper_slavin}. 

\section{Conclusion}

The paper describe the problem of stopping the process of text line recognition in a video stream. Previously presented stopping methods were described and their properties analyzed. A method based on modelling of the next integrated result is described and applied to the model of text recognition result as an alternatives matrix with extended per-character classification results. The applicability of the stopping method in these conditions is shown, and the comparative evaluation is performed against previously published methods. It was shown that the next integrated result modelling method outperforms the previously published clustering methods, even in their best achievable configurations.

\acknowledgements

This work is partially financially supported by Russian Foundation for Basic Research (projects 17-29-03170 and 19-29-09055). 

\bibliographystyle{spiebib}
\bibliography{bibliography}


\clearpage

%

\end{document}